\RequirePackage{fixltx2e}
\documentclass[11pt,english]{article}
\usepackage{mathptmx}
\usepackage[T1]{fontenc}
\usepackage[utf8]{inputenc}
\usepackage[a4paper]{geometry}
\usepackage{fancyhdr}
\usepackage{babel}
\usepackage{amsmath}
\usepackage{amssymb}
\usepackage{graphicx}
\usepackage{setspace}
\usepackage[authoryear]{natbib}
\usepackage[unicode=true,pdfusetitle,
 bookmarks=true,bookmarksnumbered=false,bookmarksopen=false,
 breaklinks=true,pdfborder={0 0 0},pdfborderstyle={},backref=false,colorlinks=false]
 {hyperref}
\hypersetup{
 linkcolor=blue,citecolor=blue,filecolor=blue,urlcolor=blue}

\makeatletter

\providecommand{\tabularnewline}{\\}

\usepackage{pstricks}
\usepackage{epsfig}
\usepackage{multicol} 
\usepackage{caption}
\usepackage{multirow}
\usepackage{tikz}
\usepackage{qtree}
\usepackage{needspace}

\usepackage{ucs}
\usepackage{tikz-qtree}
\usetikzlibrary{trees}
\usetikzlibrary{calc}

\usepackage{etoolbox}
\usepackage{ragged2e}
\raggedcolumns

\tikzset{square arrow above/.style={to path={-- ++(0,.25) -| (\tikztotarget)}}}
\tikzset{square arrow below/.style={to path={-- ++(0,-.25) -| (\tikztotarget)}}}

\let\OLDthebibliography\thebibliography
\renewcommand\thebibliography[1]{
  \OLDthebibliography{#1}
  \setlength{\parskip}{0pt}
  \setlength{\itemsep}{0pt plus 0.3ex}
}

\let\oldmaketitle\maketitle
\renewcommand{\maketitle}{%
  \oldmaketitle
  \thispagestyle{fancy}
}

\newenvironment{Figure}
  {\par\medskip\noindent\minipage{\linewidth}}
  {\endminipage\par\medskip}

\makeatother

\begin{document}
\title{Extending dependencies to the \emph{taggedPBC}: Word order in transitive
clauses}
\author{Hiram Ring\\
NTU Singapore}
\maketitle
\begin{abstract}
The \emph{taggedPBC} (\citealt{Ring:2025aa}) contains more than 1,800
sentences of pos-tagged parallel text data from over 1,500 languages,
representing 133 language families and 111 isolates. While this dwarfs
previously available resources, and the POS tags achieve decent accuracy,
allowing for predictive crosslinguistic insights (\citealt{Ring:2025ab}),
the dataset was not initially annotated for dependencies. This paper
reports on a CoNLLU-formatted version of the dataset which transfers
dependency information along with POS tags to all languages in the
\emph{taggedPBC}. Although there are various concerns regarding the
quality of the tags and the dependencies, word order information derived
from this dataset regarding the position of arguments and predicates
in transitive clauses correlates with expert determinations of word
order in three typological databases (WALS, Grambank, Autotyp). This
highlights the usefulness of corpus-based typological approaches (as
per \citealt{Baylor:2023aa,Bjerva:2024aa}) for extending comparisons
of discrete linguistic categories, and suggests that important insights
can be gained even from noisy data, given sufficient annotation. The
dependency-annotated corpora are also made available for research
and collaboration via GitHub.
\end{abstract}

\section{Introduction}

An important aspect of research on the more than 7,000 languages in
the world regards the ways in which these languages are both similar
and different. Clarifying such similarities and differences has been
the goal of linguistic typology, with languages being identified and
categorized in various ways by a variety of databases, most notably
WALS (\citealt{Dryer:2013ab}),\footnote{https://wals.info/} Grambank
\citep{Skirgard:2023ab},\footnote{https://grambank.clld.org/} Autotyp
\citep{Bickel:2023aa},\footnote{https://github.com/autotyp/autotyp-data}
Glottolog (\citealt{Hammarstrom:2024aa}),\footnote{https://www.glottolog.org}
and Lexibank (\citealt{List:2022aa}).\footnote{https://lexibank.clld.org/}
Such resources provide important information \emph{about} languages,
but often lack coverage for particular typological features and/or
languages. Additionally, the categorical determinations made by such
datasets may not apply equally to all languages, and some classifications
(word order, for example) risk pigeonhole-ing a language, as the particular
feature may be marginal or debated by specialists for particular languages.

As noted in \citet{Ring:2025aa}, a recent development has been the
creation of annotated databases that address some of these issues
by developing robust crosslinguistic corpora, most notably the Universal
Dependencies Treebanks project (UDT; \citealt{Zeman:2024ab}).\footnote{https://universaldependencies.org}
This database contains corpora from individual languages with manual
annotations for parts of speech and dependencies using a common framework.
Such a dataset can enable the automatic identification of properties
that can categorize or classify languages in various types, reflecting
an increasing interest in computational typology (see \citealt{Baylor:2023aa,Baylor:2024aa,Bjerva:2024aa}),
and supporting different typological approaches (see \citealt{Levshina:2023ab}).
Despite the encouraging nature of developments like UD, such datasets
suffer many of the same issues regarding coverage and degree of annotation
as typological databases.

The \emph{taggedPBC} was developed using computational tools to attempt
to address some of these issues. As noted by \citet{Ring:2025aa}:
\begin{quote}
With data sourced from the Parallel Bible Corpus (PBC; \citealt{Mayer:2014aa}),
computational word alignment tools allow for automatic tagging of
word classes in 1,597 languages (roughly 20\% of the world's languages)
that represent 133 families and 111 isolates. The automated part of
speech tagging method... was validated by comparing the results with
output from existing automatic taggers for high-resource languages
(SpaCy, Trankit) as well as with gold-standard tagged data in the
UDT. ... Individual corpora in the majority of these languages (over
1,500) contain more than 1,800 parallel sentences/verses, and the
smallest corpus contains more than 700 sentences/verses.
\end{quote}
This dataset represents the largest publicly available tagged crosslinguistic
database of parallel corpora, and establishes validity through correlations
with word order classfications in well-known typological databases.
As identified by Ring's study, a corpus measure derived from this
dataset (the `N1 ratio') not only distinguishes between languages
with \emph{SV}, \emph{VS}, and \emph{free} basic word order in WALS,
Grambank, and Autotyp, but also allows for accurate classification
of the intransitive word order of previously unclassified languages.
Further, in a separate study, \citet{Ring:2025ab} shows that additional
measures derived from this dataset (average lengths of nouns and verbs)
can separately classify basic intransitive word order in languages.
This measure was then tested on hand-tagged corpora not part of the
\emph{taggedPBC}, and it was found that the measure allows for accurate
classification of languages as \emph{SV} or \emph{VS} even for historical
varieties. Importantly, Ring's studies identify trends in the \emph{taggedPBC}
that are then validated on gold-standard (hand-tagged) datasets.

While Ring's studies highlight the potential for datasets like the
\emph{taggedPBC} to help answer crosslinguistic questions about language,
there are still many concerns that need to be addressed with regard
to annotation. One major concern is the quality and coverage of the
part of speech tags that can be identified via cross-lingual transfer.
This is reflected by two related concerns, namely whether the POS
tags accurately represent the scope of tags in the language, and whether
relevant tags accurately transfer between English and each specific
language in the dataset.

The first related concern can be illustrated by the simple observation
that not all languages contain all grammatical categories. While all
languages do seem to contain word classes corresponding to `nouns'
and `verbs', it is much more debatable whether all languages have
`adjectives' or `adverbs'. Further, other word classes simply don't
exist in particular languages. For example, while some languages like
Chinese require a special `classifier' word class in counting expressions,
this grammatical category has no real counterpart in English.

This leads to the second related concern, highlighting that cross-lingual
tag transfer from English, as followed by the methodology that created
the \emph{taggedPBC}, will be more successful for some word classes
than others. \citet{Ring:2025aa} does show that the methodology is
quite successful for nouns and verbs, but presumably other tags would
not transfer as well, and some (particularly those not shared by the
source and target languages) would not transfer at all. In the case
of Ring's studies, it seems that one of the reasons trends may be
observable (and can be validated on other datasets) is simply that
the quantity of observations is quite high, which mitigates (at least
to some degree) the noisy nature of the data.

These two concerns have led to particular design decisions in databases
such as the UDT, whereby certain tags without crosslinguistic counterparts
are subsumed under other categories. For the purposes of comparison,
it is necessary to ensure that comparanda share tags, which leads
to multiple levels of annotation, such that language-specific parts
of speech are annotated alongside more general categories. While there
are still issues with this approach (for example, the definition of
language-specific word classes differing widely between some languages),
it does allow for comparison of typological features at a more fine-grained
level than reported by the \emph{taggedPBC}, particularly by striving
to identify the dependency relations between words in a language.

The current paper reports on an update to the \emph{taggedPBC}\footnote{https://github.com/lingdoc/taggedPBC\label{fn:linked-repo}}
that attempts to address some of its shortcomings by extending dependency
annotations from English to each language in the dataset and converting
each dataset to CoNLL-U format.\footnote{https://universaldependencies.org/format.html}
The general idea here is that since these are parallel corpora, the
dependency relations should theoretically be maintained across translation
equivalents within corresponding sentences/verses. The following sections
lay out the methodology used to annotate dependencies (section \ref{sec:Methodology})
and show that measures derived from these corpora correspond with
basic word order determinations for transitive clauses in typological
reference databases (section \ref{sec:Correspondences}). I conclude
with some general observations and possible directions toward improving
annotations in this dataset for additional crosslinguistic investigations
(section \ref{sec:Conclusion}).

\section{Methodology\label{sec:Methodology}}

Since transferring POS tags from high-resource languages to low-resource
languages has been shown to work relatively well for common classes
such as nouns and verbs (see \citealt{Ring:2025aa} and references
therein), I use a similar methodology to transfer dependency information
across parallel sentences. With the same language-specific tokenizers,
I use the IBM Model 2 word alignment models trained by \citet{Ring:2025aa}
to identify translation equivalents in parallel verses. Since SpaCy
also has a parser for English, I then transfer dependency information
and morphological information from the English (target) sentences/verses
as processed by SpaCy to the corresponding words in the parallel verse
of the source language. Specifics of this process are contained in
the relevant code at the linked repository (footnote \ref{fn:linked-repo}).

This process allows for automatic transfer of dependencies, provided
that translation equivalents are accurately identified. For example,
in the following verse from Matthew 8:1 (Table \ref{tab:Irish-verse-tagged}),
the Irish (ISO 639-3: gle) text can be seen annotated with English
(ISO 639-3: eng) information. While there are clearly inaccuracies
in some of the transferred information, it is striking that the verb
and relevant arguments are accurately identified despite the differences
in word order (Irish: VSO, English: SVO) between the two languages.\footnote{For the sake of space, additional annotations present in the corpora
are not included here.}

\begin{table}[h]
\begin{tabular}{lllll}
\multicolumn{5}{l}{\# sent\_id = 40008001}\tabularnewline
\multicolumn{5}{l}{\# ref\_id = Matthew 8:1}\tabularnewline
\multicolumn{5}{l}{\# eng\_text = When he came down from the mountain , great multitudes
followed him .}\tabularnewline
\multicolumn{5}{l}{\# gle\_text = Tháinig sé anuas ón sliabh agus lean sluaite móra é
.}\tabularnewline
\textbf{\emph{id}} & \textbf{\emph{word}} & \textbf{\emph{pos}} & \textbf{\emph{deprel}} & \textbf{\emph{gloss\_eng}}\tabularnewline
1 & Tháinig & VERB & advcl & gloss=came\tabularnewline
2 & sé & unk & \_ & gloss=He\tabularnewline
3 & anuas & ADP & prep & gloss=down\tabularnewline
4 & ón & unk & \_ & gloss=instantly\tabularnewline
5 & sliabh & NOUN & pobj & gloss=mountainside\tabularnewline
6 & agus & unk & \_ & gloss=and\tabularnewline
7 & lean & VERB & root & gloss=followed\tabularnewline
8 & sluaite & NOUN & nsubj & gloss=crowds\tabularnewline
9 & móra & ADJ & amod & gloss=Large\tabularnewline
10 & é & unk & \_ & gloss=it\tabularnewline
11 & . & PUNCT & punct & \tabularnewline
\end{tabular}

\caption{Irish verse tagged for dependencies via English\label{tab:Irish-verse-tagged}}

\end{table}

This example shows, first, that it is possible to transfer such dependency
information between languages with parallel corpora. At the same time,
it shows that such automated transfer introduces a good bit of noise,
with corpora for certain languages being likely to contain more noise
than others, depending on how closely it aligns with English on various
linguistic properties. As with the previous observations regarding
noise in the original \emph{taggedPBC} dataset, it may still be possible
to observe crosslinguistic trends across languages.

As \citet{Ring:2025aa} notes, it is incumbent on researchers in computational
typology to demonstrate validity of the various measures used as well
as the validity of the underlying data. In the case of POS-tagging,
\citet{Ring:2025aa} validates the accuracy of the POS tags transferred
from English by comparing with SOTA taggers and hand-tagged corpora
for a large number of (non-English) languages. In the case of dependency
relations, there are a few issues with validating the accuracy of
the transferred dependencies. Since dependency relations are less
``fixed'' in some senses than word classes, such that the same word
may occur in multiple dependency relations in different clauses, it
is more difficult to assess their accuracy. This would require a parallel
dataset annotated for dependency relations, or that a portion of the
current dataset be hand-annotated for dependency relations. Another
possibility would be to assess against automated parsers such as developed
by \citet{Ustun:2020aa,Choudhary:2023aa}.

Yet another possibility, however, is to assess how well various measures
derived via dependencies align with expert typological determinations.
In this sense, we skip a narrow validation in favor of a broader validation
with known/expected outcomes. The correlation of measures extracted
from this dataset with existing expert classifications is what we
turn to in the following section.

\section{Correspondences with transitive word order\label{sec:Correspondences}}

The CoNLL-U formatted version of the \emph{taggedPBC} includes quite
a bit of additional information that the original pos-tagged dataset
did not. While the original tagged dataset was able to show a distinction
between \emph{SV}, \emph{VS}, and \emph{free} word order languages
on the basis of the N1 ratio, the study focused on differentiating
between languages on the basis of word order associated with intransitive
verbs, which take a single (``Subject'') argument. In contrast,
the dependencies transferred into the the CoNLL-U formatted version
additionally encode ``Subject'' and ``Object'' information for
arguments, allowing for a 6-way distinction in sentence/clause structure
(\emph{VSO}, \emph{VOS}, \emph{SVO}, \emph{OVS}, \emph{SOV}, \emph{OSV})
for transitive verbs.

It should be noted here that the existence of ``Subject'' and ``Object''
relations in all languages is somewhat debatable. This is because
``subject'' often refers to a particular alignment in a language
between arguments of a transitive verb/clause and arguments of an
intransitive verb/clause. The ``subject'' in a Nominative-Accusative
language is taken to be the argument of a transitive clause that is
marked (in some manner) similarly to the single argument of an intransitive
clause. This generally corresponds to the ``agent'' of the verbal
event (i.e. ``he hits her'' {[}tr{]} vs ``he runs'' {[}intr{]}).
In contrast, an Ergative-Absolutive language marks the single argument
of an intransitive clause the same as the ``undergoer'' or ``recipient''
argument of a transitive clause (the ``Object''). However, since
the UD framework does not make this distinction at the crosslinguistic
level (leaving it to language-specific annotations to clarify) I adopt
the ``Subject'' and ``Object'' terminology with the expectation
that future annotation efforts can address this important point.

\begin{figure}[h]
\includegraphics[width=0.7\columnwidth]{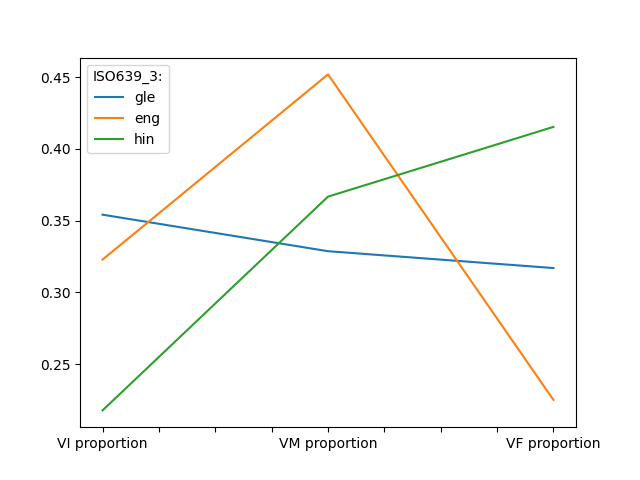}

\caption{Word order proportions for 3 languages\label{fig:Word-order-proportions}}

\end{figure}

Despite such potential ambiguities and errors in annotation, we can
follow \citet{Ring:2025aa}'s process for deriving the N1 ratio in
an attempt to derive a corpus-based measure of transitive word order. To do this,
for each language we simply count the number of verses where
both the dependency relations ``subject'' and ``object'' are identified
along with a word tagged as ``VERB''. For each of the six transitive
word order patterns, we can then derive a ratio (or percentage/proportion)
by dividing the number of occurrences of that pattern by the total
number of sentences/verses in the corpus that contain all three elements,
allowing for gradient observations as recommended by \citet{Levshina:2023ab}.
As an example, Figure \ref{fig:Word-order-proportions} plots word
order proportions for three languages with different word orders:
Irish (VSO; gle), English (SVO; eng), and Hindi (SOV; hin).

That such gradient observations can be derived from a corpus are interesting,
but the real question is whether they are meaningful. To test this,
I extracted transitive word order information from existing typological
databases (WALS, Grambank, Autotyp) and observed correlations with
these expert determinations and the percentages derived from the CoNLL-U
\emph{taggedPBC} corpora. The observations present in these databases
are not without comparability issues, however. For example, while
WALS identifies individual languages as having one of the six basic
transitive word orders (or ``free'' word order), Grambank only identifies
whether one of three verb positions (verb-initial, verb-medial, verb-final)
are attested as well as whether the word order is ``fixed'' or ``free''.
Further, while Autotyp identifies basic transitive word order similarly
to WALS (replacing ``S'' with ``A''), some languages are identified
by the position of the verb in relation to other elements (Tohono
O'odham {[}ood{]} is listed as ``Vxx'', for example). Accordingly,
in the interest of comparibility I reduced the 7 transitive word order
possibilities to 4: Verb initial (``VI'', subsuming VSO and VOS),
Verb medial (``VM'', subsuming SVO and OVS), Verb final (``VF'',
subsuming SOV and OSV), and ``free''.

Combining the three typological databases allows us to compare word
order classifications with 961 languages that are also present in
the \emph{taggedPBC}. In Figures \ref{fig:Corpus-derived-Verb-initial-prop}-\ref{fig:Corpus-derived-Verb-final-propor}
we can observe some general trends regarding proportions of each observed
word order pattern in the corpus according to their typological classification.
Interestingly, one-way ANOVAs (see stats at the linked repository,
fn \ref{fn:linked-repo}) show that some of the word order proportions
meaningfully differentiate between certain classifications, but not
others. In particular, ``free'' word order languages are extremely
difficult to distinguish from languages classified as having other
basic word orders. This makes sense given that ``free'' word order
languages allow any variation of the six attested patterns, meaning
that some of their corpora would show proportions aligning with VI
languages while others would show proportions aligning with VM or
VF languages.

\begin{multicols}{2}
\begin{Figure}
 \centering\includegraphics[width=0.9\columnwidth]{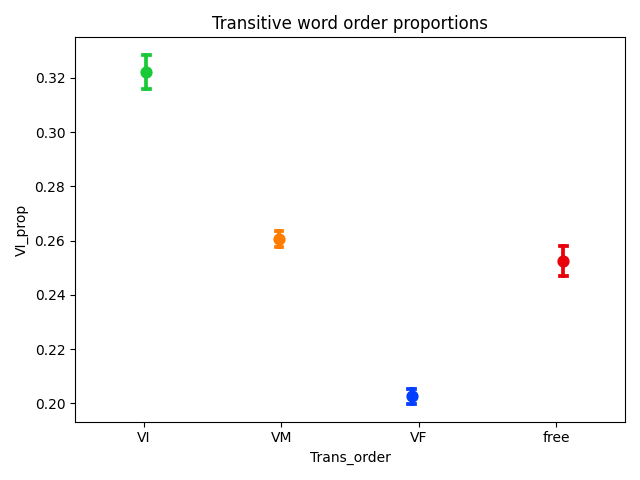}

  \captionof{figure}{Verb-initial proportions}\label{fig:Corpus-derived-Verb-initial-prop}\end{Figure}
\begin{Figure}
 \centering\includegraphics[width=0.9\columnwidth]{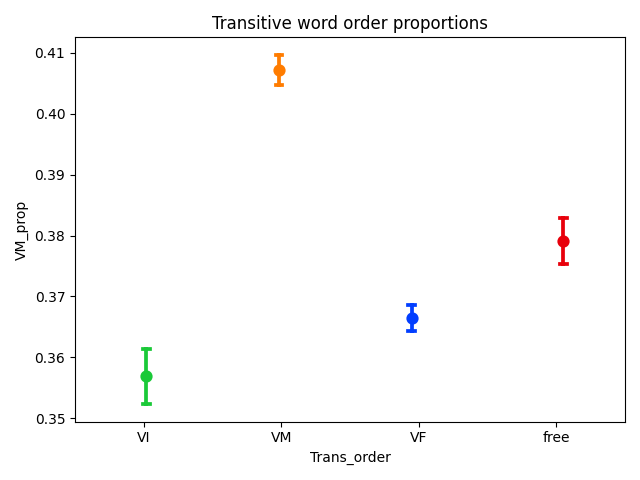}

  \captionof{figure}{Verb-medial proportions}\label{fig:Corpus-derived-Verb-medial-propo}\end{Figure}
\begin{Figure}
 \centering\includegraphics[width=0.9\columnwidth]{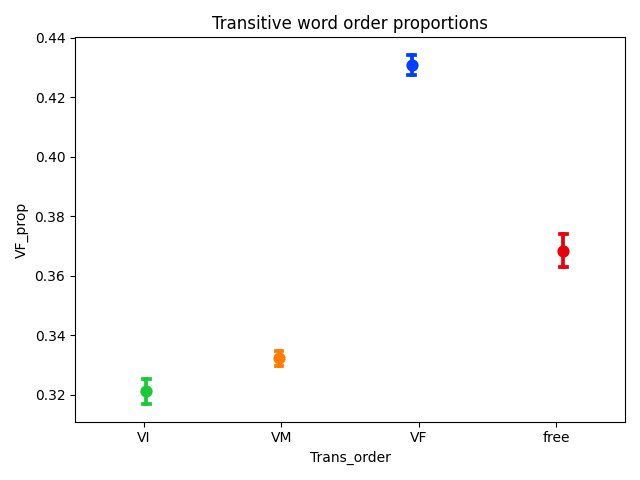}

  \captionof{figure}{Verb-final proportions}\label{fig:Corpus-derived-Verb-final-propor}

\end{Figure}
\begin{Figure}
 \centering\includegraphics[width=0.9\columnwidth]{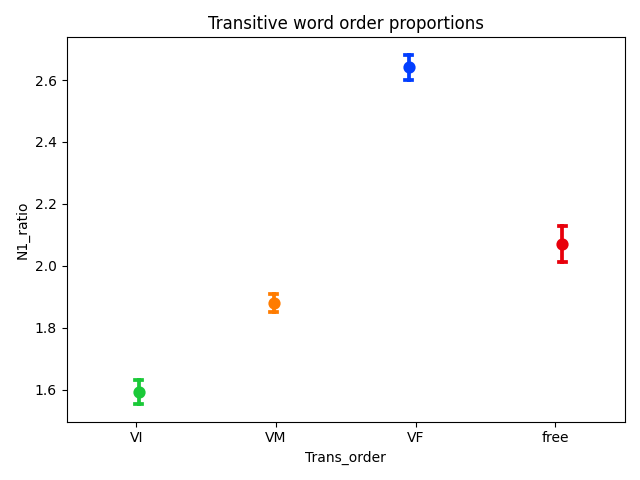}

\noindent   \captionof{figure}{N1 ratio correspondences}\label{fig:Corpus-derived-N1-ratio}  \end{Figure}
\end{multicols}

Unfortunately, this means that it would be very difficult to determine
whether languages have a ``fixed'' or ``free'' word order based
simply on proportions of word order occurrences in corpora. Once ``free''
word order languages are separated from ``fixed'' word order languages,
it then seems possible to classify the ``fixed'' languages as belonging
to one of the 3 remaining categories. Yet identifying which are ``free''
and which are ``fixed'' does not seem to be possible from proportions
to begin with, making this effort moot.

However, exploratory analysis reveals another correlation. Conducting
an ANOVA using the N1 ratio as the dependent variable and the VI,
VM, VF and free classifications as the fixed factors reveals that
the N1 ratio clearly differentiates between these groups (p < 0.003;
see Figure \ref{fig:Corpus-derived-N1-ratio}). Surprisingly, this
measure, derived simply from a count of verses that start with an
argument divided by those that start with a predicate, allow us to
determine whether a language can be classified as having VI, VM, VF,
or free word order, showing a clear alignment with existing typological
databases.

\section{Conclusion\label{sec:Conclusion}}

The preceding sections have shown that an updated version of the \emph{taggedPBC}
which transfers dependency information from English aligns quite well
with existing classfications for (transitive) word order. While this
dataset is still quite noisy, it suggests that large datasets created
with the aid of automatic annotation can support crosslinguistic investigations
of coarse features such as word order. Interestingly, via crosslingual
transfer, all languages in the dataset contain sentences/verses with
``subject'', ``object'', and ``verb''. Further, all languages
have at least one sentence/verse labeled for each of the six transitive
word order patterns.

The word order investigations presented here highlight both the importance
of a gradient approach and the possibility of identification of discrete
categories for word order. Observation of corpora indicates that all
languages show all word orders to differing degrees. At the same time,
the significant correlation between the N1 ratio and expert judgements
in typological databases highlights the validity of such categories.
Some languages clearly prioritize (or grammaticalize) one order over
another, and this is reflected in corpus statistics.

There are still many concerns to address with a database of this type,
however. Although the results of the investigations reported here
indicate that the current dataset is sufficient for coarse-grained
analysis, a degree of noise has been introduced via the current methodology.
For example, POS tags and dependency relations are all English-centric,
with some language-specific tags (such as numeral classifiers) largely
or completely missing from the dataset. Additionally, none of the
languages in this dataset have morphological annotations, since the
methodology used depends on word alignments rather than morphological
alignment. Here, we are at the mercy of word tokenization via whitespace
based on tokenization conventions of individual languages. Further,
while an attempt has been made to identify a particular word's HEAD
relation,\footnote{In the sense used by the UD CoNLL-U format, under field \#7: https://universaldependencies.org/format.html}
this is a problematic relation to identify automatically on the basis
of translation, which means that dependent clauses are not distinguished
from main clauses, a concern that might make quite a bit of difference
for identification of such things as word order patterns.

Each of these concerns can be addressed, perhaps gradually and progressively,
through manual annotation as well as new methodologies. To consider
some possibilities, it is likely that existing POS-tagging can be
improved for particular languages by leveraging hand-tagged ``gold
standard'' corpora to train better taggers, and the dataset itself
could be expanded by incorporating hand-tagged data from other genres.
Additionally, while manual annotation from experts on particular languages
could go a long way toward improving not only POS tags but also dependency
relations, the nature of the current dataset as being parallel corpora
also lends itself to support from non-expert annotators. Since dependency
information is expected to align in parallel verses/sentences, if
translation equivalents of particular words can be identified by annotators
then the relevant dependencies can also be annotated (with a high
degree of certainty), without the need for detailed knowledge of a
particular language.

In terms of new methodologies, one example is the possibility of training
a ``universal tagger'' by hand-tagging a select group of languages
that represent the diversity of typological features and word classes
identified in typological databases. Once a complete set of word classes
is identified for languages of the world, hand-tagging a portion of
relevant language datasets could allow for the observation of patterns
that correlate with the distribution of particular tags and related
word classes, similar to existing approaches to crosslinguistic tag
transfer (i.e. \citealt{Imani:2022aa}). Having statistical priors
that correlate with language-specific tags could allow a computational
model to predict or identify tags for specific languages (such as
a ``numeral classifier'') in one-shot situations where the actual
tags are unknown. Similar improvements could be made for existing
multilingual approaches to tagging dependency relations (\citealt{Choudhary:2023aa}).

Overall this dataset represents a sea change in how computational
typology (and other kinds of crosslinguistic investigations) can be
conducted. While there are many potential improvements that should
be made to the data itself, such long-term goals are balanced by our
ability to observe coarse-grained trends across many languages. With
the aid of computational tools and statistical methods, there is much
we can learn about human language through comparison of parallel data
of this kind. Although these findings may change somewhat as the underlying
data is more fully developed, I have shown here that features found
in the data correlate quite well with word order observations for transitive clauses in
languages of the world, supporting the claim that corpus-derived features
are a valuable asset in crosslinguistic investigations.

\bibliographystyle{linguistics}
\phantomsection\addcontentsline{toc}{section}{\refname}\bibliography{my-bibliog}

\begin{thebibliography}{16}
\providecommand{\natexlab}[1]{#1}
\providecommand{\url}[1]{\texttt{#1}}
\providecommand{\urlprefix}{URL }
\expandafter\ifx\csname urlstyle\endcsname\relax
  \providecommand{\doi}[1]{doi:\discretionary{}{}{}#1}\else
  \providecommand{\doi}{doi:\discretionary{}{}{}\begingroup
  \urlstyle{rm}\Url}\fi
\providecommand{\eprint}[2][]{\url{#2}}

\bibitem[{Baylor et~al.(2023)Baylor, Ploeger, \& Bjerva}]{Baylor:2023aa}
Baylor, Emi, Esther Ploeger, \& Johannes Bjerva. 2023.
\newblock The past, present, and future of typological databases in {NLP}.
\newblock In Bouamor, Houda, Juan Pino, \& Kalika Bali (eds.), \emph{Findings
  of the Association for Computational Linguistics: EMNLP 2023}, pp.
  1163--1169. Singapore: Association for Computational Linguistics.
\newblock \urlprefix\url{https://aclanthology.org/2023.findings-emnlp.82/}.

\bibitem[{Baylor et~al.(2024)Baylor, Ploeger, \& Bjerva}]{Baylor:2024aa}
Baylor, Emi, Esther Ploeger, \& Johannes Bjerva. 2024.
\newblock Multilingual gradient word-order typology from universal
  dependencies.
\newblock \urlprefix\url{https://arxiv.org/abs/2402.01513}.
\newblock \eprint{2402.01513}.

\bibitem[{Bickel et~al.(2023)Bickel, Nichols, Zakharko, Witzlack-Makarevich,
  Hildebrandt, Rie{\ss}ler, Bierkandt, Z{\'u}{\~n}iga, \& Lowe}]{Bickel:2023aa}
Bickel, Balthasar, Johanna Nichols, Taras Zakharko, Alena Witzlack-Makarevich,
  Kristine Hildebrandt, Michael Rie{\ss}ler, Lennart Bierkandt, Fernando
  Z{\'u}{\~n}iga, \& John~B Lowe. 2023.
\newblock The {AUTOTYP} database (v1.1.1).

\bibitem[{Bjerva(2024)}]{Bjerva:2024aa}
Bjerva, Johannes. 2024.
\newblock The role of typological feature prediction in nlp and linguistics.
\newblock \emph{Computational Linguistics}, 50(2):781--794.
\newblock ISSN 0891-2017.
\newblock \urlprefix\url{https://doi.org/10.1162/coli\_a\_00498}.
\newblock
  \eprint{https://direct.mit.edu/coli/article-pdf/50/2/781/2457439/coli\_a\_00498.pdf}.

\bibitem[{Choudhary \& O{'}riordan(2023)}]{Choudhary:2023aa}
Choudhary, Chinmay \& Colm O{'}riordan. 2023.
\newblock Multilingual end-to-end dependency parsing with linguistic typology
  knowledge.
\newblock In Beinborn, Lisa, Koustava Goswami, Saliha Murado{\u{g}}lu, Alexey
  Sorokin, Ritesh Kumar, Andreas Shcherbakov, Edoardo~M. Ponti, Ryan Cotterell,
  \& Ekaterina Vylomova (eds.), \emph{Proceedings of the 5th Workshop on
  Research in Computational Linguistic Typology and Multilingual NLP}, pp.
  12--21. Dubrovnik, Croatia: Association for Computational Linguistics.
\newblock \urlprefix\url{https://aclanthology.org/2023.sigtyp-1.2/}.

\bibitem[{Dryer \& Haspelmath(2013)}]{Dryer:2013ab}
Dryer, Matthew~S. \& Martin Haspelmath (eds.). 2013.
\newblock \emph{WALS Online (v2020.4)}.
\newblock Zenodo.

\bibitem[{Hammarstr{\"o}m et~al.(2024)Hammarstr{\"o}m, Forkel, Haspelmath, \&
  Bank}]{Hammarstrom:2024aa}
Hammarstr{\"o}m, Harald, Robert Forkel, Martin Haspelmath, \& Sebastian Bank.
  2024.
\newblock \emph{Glottolog 5.1}.
\newblock Leipzig: Max Planck Institute for Evolutionary Anthropology.
\newblock \urlprefix\url{http://glottolog.org}.
\newblock Accessed on 2025-04-03.

\bibitem[{Imani et~al.(2022)Imani, Severini, Jalili~Sabet, Yvon, \&
  Sch{\"u}tze}]{Imani:2022aa}
Imani, Ayyoob, Silvia Severini, Masoud Jalili~Sabet, Fran{\c{c}}ois Yvon, \&
  Hinrich Sch{\"u}tze. 2022.
\newblock Graph-based multilingual label propagation for low-resource
  part-of-speech tagging.
\newblock In Goldberg, Yoav, Zornitsa Kozareva, \& Yue Zhang (eds.),
  \emph{Proceedings of the 2022 Conference on Empirical Methods in Natural
  Language Processing}, pp. 1577--1589. Abu Dhabi, United Arab Emirates:
  Association for Computational Linguistics.

\bibitem[{Levshina et~al.(2023)Levshina, Namboodiripad, Allassonni{\`e}re-Tang,
  Kramer, Talamo, Verkerk, Wilmoth, Rodriguez, Gupton, Kidd, Liu, Naccarato,
  Nordlinger, Panova, \& Stoynova}]{Levshina:2023ab}
Levshina, Natalia, Savithry Namboodiripad, Marc Allassonni{\`e}re-Tang, Mathew
  Kramer, Luigi Talamo, Annemarie Verkerk, Sasha Wilmoth, Gabriela~Garrido
  Rodriguez, Timothy~Michael Gupton, Evan Kidd, Zoey Liu, Chiara Naccarato,
  Rachel Nordlinger, Anastasia Panova, \& Natalia Stoynova. 2023.
\newblock Why we need a gradient approach to word order.
\newblock \emph{Linguistics}, 61(4):825--883.
\newblock \urlprefix\url{https://doi.org/10.1515/ling-2021-0098}.

\bibitem[{List et~al.(2022)List, Forkel, Greenhill, Rzymski, Englisch, \&
  Gray}]{List:2022aa}
List, Johann-Mattis, Robert Forkel, Simon~J. Greenhill, Christoph Rzymski,
  Johannes Englisch, \& Russell~D. Gray. 2022.
\newblock Lexibank, a public repository of standardized wordlists with computed
  phonological and lexical features.
\newblock \emph{Scientific Data}, 9(1):316.
\newblock \urlprefix\url{https://doi.org/10.1038/s41597-022-01432-0}.

\bibitem[{Mayer \& Cysouw(2014)}]{Mayer:2014aa}
Mayer, Thomas \& Michael Cysouw. 2014.
\newblock Creating a massively parallel {B}ible corpus.
\newblock In \emph{Proceedings of The International Conference on Language
  Resources and Evaluation (LREC)}, pp. 3158--3163. Reykjavik.

\bibitem[{Ring(2025{\natexlab{a}})}]{Ring:2025aa}
Ring, Hiram. 2025{\natexlab{a}}.
\newblock The \emph{taggedPBC}: Annotating a massive parallel corpus for
  crosslinguistic investigations.
\newblock \urlprefix\url{https://arxiv.org/abs/2505.12560}.
\newblock \eprint{2505.12560}.

\bibitem[{Ring(2025{\natexlab{b}})}]{Ring:2025ab}
Ring, Hiram. 2025{\natexlab{b}}.
\newblock Word length predicts word order: "min-max"-ing drives language
  evolution.
\newblock \urlprefix\url{https://arxiv.org/abs/2505.13913}.
\newblock \eprint{2505.13913}.

\bibitem[{Skirg{\aa}rd et~al.(2023)Skirg{\aa}rd, Haynie, Blasi,
  Hammarstr{\"o}m, \& et~al}]{Skirgard:2023ab}
Skirg{\aa}rd, Hedvig, Hannah~J. Haynie, Dami{\'a}n~E. Blasi, Harald
  Hammarstr{\"o}m, \& et~al. 2023.
\newblock Grambank reveals global patterns in the structural diversity of the
  world's languages.
\newblock \emph{Science Advances}, 9.

\bibitem[{{\"U}st{\"u}n et~al.(2020){\"U}st{\"u}n, Bisazza, Bouma, \& van
  Noord}]{Ustun:2020aa}
{\"U}st{\"u}n, Ahmet, Arianna Bisazza, Gosse Bouma, \& Gertjan van Noord. 2020.
\newblock {UD}apter: Language adaptation for truly {U}niversal {D}ependency
  parsing.
\newblock In \emph{Proceedings of the 2020 Conference on Empirical Methods in
  Natural Language Processing (EMNLP)}, pp. 2302--2315. Online: Association for
  Computational Linguistics.
\newblock \urlprefix\url{https://www.aclweb.org/anthology/2020.emnlp-main.180}.

\bibitem[{Zeman et~al.(2024)Zeman, Nivre, Abrams, \& et~al}]{Zeman:2024ab}
Zeman, Daniel, Joakim Nivre, Mitchell Abrams, \& et~al. 2024.
\newblock Universal dependencies 2.14.
\newblock {LINDAT}/{CLARIAH}-{CZ} digital library at the Institute of Formal
  and Applied Linguistics ({{\'U}FAL}), Faculty of Mathematics and Physics,
  Charles University.

\end{thebibliography}

\end{document}